\documentclass[runningheads]{llncs}
\usepackage{graphicx}
\usepackage{multirow}
\usepackage{tcolorbox}
\usepackage{amsmath}
\begin{document}
\title{Faster feature selection with a Dropping Forward-Backward algorithm}
%
%\titlerunning{Abbreviated paper title}
% If the paper title is too long for the running head, you can set
% an abbreviated paper title here
%
\author{Thu Nguyen}%\inst{1}\orcidID{0000-0001-7044-1731}}
\authorrunning{Thu Nguyen}
% First names are abbreviated in the running head.
% If there are more than two authors, 'et al.' is used.
%
\institute{University of Louisiana at Lafayette, Lafayette LA 70504, USA} 
%\email{thu.nguyen@louisiana.edu}}
%\url{http://www.springer.com/gp/computer-science/lncs} \and
%ABC Institute, Rupert-Karls-University Heidelberg, Heidelberg, Germany\\
%\email{\{abc,lncs\}@uni-heidelberg.de}}
%
\maketitle              % typeset the header of the contribution

\begin{abstract}
 In this era of big data, feature selection techniques, which have long been proven to simplify the model, makes the model more comprehensible, speed up the process of learning, have become more and more important. Among many developed methods, forward and stepwise feature selection regression remained widely used due to their simplicity and efficiency. However, they all involving rescanning all the un-selected features again and again. Moreover, many times, the backward steps in stepwise deem unnecessary, as we will illustrate in our example. These remarks motivate us to introduce a novel algorithm that may boost the speed up to 65.77\%  compared to the stepwise procedure while maintaining good performance in terms of the number of selected features and error rates. Also, our experiments illustrate that feature selection procedures may be a better choice for high-dimensional problems where the number of features highly exceeds the number of samples.
\end{abstract}
\section{Introduction}
In this era of big data, the growth of data poses challenges for effective data management and inference. Real-world data usually contain a lot of redundant or irrelevant features that can derail the learning performance. Moreover, for high-dimensional data, a critical issue is that the number of features highly surpasses the number of samples, which could cause the models to overfit, and performance on test data suffer. This is well known as the curse of dimensionality or the $n\gg p$ problem. To deal with this issue, various feature extraction and feature selection methods have been developed (see \cite{liu2012feature,kumar2014feature} or the related for reviews). However, the feature extraction methods create sets of new features that we can not directly interpret. Moreover, since those approaches use all the features available during training, it does not help to reduce the cost of collecting data in the future. Feature selection, on the other hand, helps to maintain the meanings of the original features, reducing the cost of storage and collecting data in the future, by removing irrelevant or redundant features. %This, of course, relies upon the central premise that the data contains some features that are irrelevant or redundant, and therefore can be removed without causing much loss for the model. 

Feature selection techniques can be classified into three categories: filter, wrapper, and embedded. The filter approaches (Markov Blanket Filtering, t-test,etc.) extract features from data without involving any learning. The wrappers methods (genetic algorithm, sequential search, etc.), on the other hand, use learning techniques to evaluate the importance of the features. Finally, the embedded approaches (random forest, least absolute shrinkage and selection operator, etc.) combine the feature selection steps with the classifier construction process.

In the wrapper approach, some of the most popular methods are forward, backward and stepwise feature selection. Forward selection starts with an empty model. Then, it sequentially adds to the model the feature that best improves the fit the most in terms of the criterion being used. This method is well known for its speed but it may select some features at some steps and later add some other features that make the inclusion of the previous ones redundant. Backward selection avoids this problem by sequentially remove the least useful feature, one at a time. However, it is computationally expensive and can only be applied when the number of samples is much larger than the number of features (see \cite{couvreur2000optimality}). Stepwise regression is a combination of these two methods. It firstly adds features to the model sequentially as in forward feature selection. In addition, after adding a new feature, it removes the features that are no longer important in the model after the inclusion of the new one. Related methods have been developed to boost the efficiency of these methods. \cite{mao2004orthogonal} incorporating Gram–Schmidt and Givens orthogonal transforms into forward and backward procedures for classification tasks, respectively. This makes the features de-correlated in the orthogonal space so that each feature can be independently evaluated and selected.
\cite{billings2007sparse} proposes a forward orthogonal algorithm, with mutual information interference for regression at the cost of doing orthogonal transforma-tions. %\cite{zhang2011adaptive} proposed a combination that takes a backward step when the squared error increase is no more than half of the squared error decrease in the prior corresponding forward step. However, there has not been any study that compares its efficiency and speed with stepwise procedure yet.  

In this paper, we introduce a novel method for feature selection that has great performance in terms of speed, error rates and number of selected features. Our contributions from this paper are of five folds: (1) We point out some deficiencies of forward and stepwise feature selection. (2) We propose a new scheme that gives much faster training time than stepwise while maintaining good results in terms of the number of selected features and error rates. (3) We demonstrate the power of our approach in regression and classification tasks using simulated and real-world data. (4) We point out that feature selection may be preferable to feature extraction in the $p\gg n$ problems. (5) We illustrate that the time it takes for a feature selection procedure to stop depends on not only the dimension of the data but also the sparsity of the resulting model.

The structure of the remaining parts of this paper is as follows. In section \ref{review}, we review the forward, and stepwise algorithms for feature selection, and point out the issues associate  with these approaches. Those serve as motivations for our approach. Next, in section \ref{dfb}, we introduce our \textit{dropping forward-backward algorithm}. Then, in section \ref{exper}, we show how powerful our approach surpasses stepwise selection and another intuitive forward-backward scheme both in terms of error reduction and the number of features selected on simulated and real datasets. Finally, in section \ref{concl}, we summarize the main points of this paper.

\section{Forward, backward, and stepwise feature selection}\label{review}

\begin{center}
	\fbox{%
		\begin{minipage}{4.7 in}
	\textbf{Forward feature selection algorithm:}	
	\begin{itemize}
		\item Input: a set of features $C=\{X_1,X_2,...,X_p\}$, response $Y$, $\alpha$-to enter value, selection criterion.
		
		\item Output: a set $R\subset C$ of relevant features.
		
		\item Procedure: 
		Sequentially add to R a feature that improves the fit the most in terms of the criterion being used.  Stop when no feature can improve the fit more than $\alpha$.		
	\end{itemize}
\end{minipage}}
\end{center}\label{Forward selection}
Forward algorithm has been used widely due to its computational efficiency, along with the possibility to deal with the $p\gg n$ problems, where the number of features highly exceeds the number of observations. However, some features included by forward steps may appear redundant after the inclusion of some other features. About the sufficient conditions for forward feature selection to recover the original model and its stability, we refer to \cite{tropp2004greed} and \cite{donoho2005stable} for further readings.
\begin{center}
\fbox{%
\begin{minipage}{4.7 in}

	\textbf{Stepwise feature selection algorithm:}	
	\begin{itemize}
		\item Input: a set of features $C=\{X_1,X_2,...,X_p\}$, response $Y$, $\beta$-to remove value, selection criterion.
		
		\item Output: a set $R\subset C$ of relevant features.
		
		\item Procedure:
		
		\begin{enumerate}
			
			\item 			
			\begin{enumerate}				
				\item \textit{Forward step: } Add to R a feature that improves the fit the most in terms of the criterion being used. 
				\item \textit{Backward step:} Sequentially remove the least useful feature in the model, one at a time, if it worsen the model no more than an amount of $\beta$, in terms of the given criterion. Stop when the removal of any feature in the model causes the fit to decrease more than $\beta$.
							
			\end{enumerate}			
			\item Stop when no feature can improve the fit more than an amount of $\alpha$.			
		\end{enumerate}		
	\end{itemize}
\end{minipage}}
\end{center}
Stepwise selection appears to be a remedy to the forward error in forward selection. It adds features to the model sequentially as in forward feature selection. In addition, after adding a new feature, this approach removes the features that are no longer important in the model. However, there is a computational cost associated with the backward steps that remove unnecessary features. Sometimes, this raises the question about how likely the forward scheme commits an error like that. \cite{zhang2011adaptive} proposes an algorithm that takes a backward step only when the squared error is no more than half of the squared error decrease in the earlier forward steps. However, this still gives rise to the same question of whether checking to take backward steps like that worth the effort. This motivates us to do some experiments to gain some insight into the problem. 
\begin{table}
	\caption{The average number of backward steps taken by stepwise procedure according to the number of features in the model.}\label{b1}	
	\begin{center}
		\begin{tabular}{|c|c|c|c|c|c|}
			\hline
			\begin{tabular}[c]{@{}c@{}}Number of features\\included in the model \end{tabular}                   & $4$& $8$ & $12$ & $16$ & $20$ \\ \hline
			\begin{tabular}[c]{@{}c@{}}Average number of backward \\ backward steps taken by \\  stepwise procedure\end{tabular} &$\;\;\;  0  \;\;\; $   & 0.013    & 0.004     & 0.003    & 0.003 \\ \hline
		\end{tabular}
	\end{center}
\end{table}

Table \ref{b1} and \ref{b2} show results from Monte Carlo simulation, with data from $80-$ dimensional multivariate normal distribution with sample size $n=80$. For the first table, we vary the number of features included in the model, repeat each experiment 1000 times, and compute the average number of backward steps taken by stepwise procedure.  For the second table, we vary the maximum correlation among features and generate correlation values for the multivariate normal distribution from 0 to the maximum value. We repeat each experiment 1000 times and compute the average number of backward steps taken by stepwise procedure. 

\begin{table}
	\caption{Average number of backward steps taken by stepwise procedure according to correlation.}\label{b2}	
	\begin{center}
		\begin{tabular}{|c|c|c|c|c|c|}
			\hline
			\begin{tabular}[c]{@{}c@{}}Maximum correlation\\ among different features\end{tabular}                   & $0.3$ & $0.4$ & $0.5$ & $0.6$ & $0.7$ \\ \hline
			\begin{tabular}[c]{@{}c@{}}Average number of  \\  backward steps taken by \\  stepwise procedure\end{tabular} & $\;\;\; 0 \;\;\; $    & $\;\;\; 0\;\;\; $     & $\;\;\; 0 \;\;\; $    &$\;\;\;  0\;\;\; $     & 0.001 \\ \hline
		\end{tabular}
	\end{center}
\end{table}
From these tables, we see that many times, the effort to check whether to take a  backward step or not does not worth the computational price. Rather, we could simply do forward feature selection to get a list R, and then do backward selection on R to correct the mistakes that forward selection scheme may have made. We shall refer to this as \textit{forward-backward algorithm}. For regression, this is reasonable, as the order of features in the model does not affect their corresponding coefficients. That can be seen directly from the following theorem:
\begin{theorem}\label{theorem}
	Suppose that we have a regression model
	\begin{equation}
		Y=X\beta + \epsilon,
	\end{equation}
	where $\epsilon \sim \mathcal{N}(0,\sigma^2I)$.
	Suppose $X=[x_1,x_2,...,x_p]$ where $x_i$ is the $i^{th}$ column vector of $X$, and $\hat{\beta} = (\hat{\beta}_1,...,\hat{\beta}_p)'$ is the least square estimate of $\beta$. Let $Z$ be the resulting matrix if we swap any two columns $x_i, x_j$ ($i<j$) of $X$. Consider the model 
	\begin{equation}
		Y=Z\gamma + \epsilon'
	\end{equation}
	then we can get the least square estimate $\widehat{\gamma}$ of $\gamma$ by swapping the $i^{th}, j^{th}$ position of the old $\hat{\beta}$, i.e.,
	\begin{equation}
		\widehat{\gamma} = (\hat{\beta}_1,...,\hat{\beta}_{i-1},\hat{\beta}_j,\hat{\beta}_{i+1},...,\hat{\beta}_{j-1},\hat{\beta}_i,\hat{\beta}_{j+1},...,\hat{\beta}_p)'.
	\end{equation}
\end{theorem}

%(Note that the order of a feature in the model does not affect its corresponding coefficient. However, the order of feature inclusion may affect whether some features are included in the model or not.)

Another point worth noticing is that all of the algorithms mentioned above require scanning over and over the remaining features in the pool when adding a new feature. This makes the algorithms suffer higher computational cost than necessary. Therefore, in the next section, we introduce a new algorithm that can remedy these inefficiencies.

\section{Feature selection by Dropping Forward-Backward algorithm} \label{dfb}
As the deficiencies of forward and stepwise feature selection algorithms are pointed out in the previous section, we introduce the following \textit{dropping forward-backward scheme} to improve these inefficiencies.

\begin{center}
	\fbox{%
		\begin{minipage}{4.7 in}
	\textbf{General dropping forward-backward scheme:}
	\begin{enumerate} 
		\item \textbf{Input}: a set of feature $C=\{X_1,X_2,...,X_p\}$, response $Y$, $\alpha$-to enter threshold, $\beta$-to remove threshold, selection criterion.
		
		\item \textbf{Output}: a set $R\subset C$ of relevant features.
		
		\item \textbf{Forward dropping steps:} Sequentially add to R the feature that improve the fit in term of the criterion being used the most. Remove from C this feature and the features that can not improve the fit more than an amount of $\beta$. Stop when no feature can improve the fit more than $\alpha$.
		
		\item \textbf{Re-forward steps:}	
		\begin{enumerate}
			\item  $C=\{X_1,X_2,...,X_p\}\setminus R$,
			
			\item Sequentially add to R a feature that best improve the fit in term of the criterion being used. Stop when no feature can improve the fit more than  $\alpha$.
			
		\end{enumerate}	
		
		\item \textbf{Backward steps:} Sequentially remove from $R$ the least useful feature, one at a time, if removing it causes the fit to decrease no more than an amount of $\beta$, until the removal of any feature in the model cause the fit to decrease more than $\beta$, in terms of the criterion being used.
		
	\end{enumerate}	
\end{minipage}}
\end{center}

Note that in the dropping forward-backward scheme above, the forward steps are very similar to the forward algorithm, except that we temporarily remove all the features in the pool that can not improve the fit more than an amount of $\beta$ in terms of the criterion being used. This helps reduce the computational cost of rescanning through the features that temporarily do not seem to be able to improve the model a lot compared to other features. Though, after that we do forward steps again in the \textit{re-forward steps}, with all the features that have not been included in the model yet, to account for the possible correlation that may improve the fit. Moreover, instead of taking a backward step after every forward move, we only take a backward step at the end of all forward steps to remove the redundant features that remained in the model. This is to correct the error that forward steps may make and avoid the computational cost of checking for a backward move after every inclusion of a new feature. 

Note that higher $\beta$ will results in more feature dropping and less re-scanning during forward dropping moves. However, depending on the data and the chosen criterion, for high dimensional data, we may prefer to use lower $\beta$. The reason is higher  $\beta$ may result in too many feature dropping, which implies that much fewer features have chances to get into the model during forward dropping moves. This causes the forward dropping moves to terminate early, and we have to re-scan a lot of features during re-forward steps.

Another worth noticing point is that after the forward dropping steps are the re-forward steps. Therefore, after the forward dropping steps, the algorithm gives ranks to the importance of the features, and it possible to specifies the maximum number of features to be included in the model in case one wishes for a smaller set of features than what the thresholds may produce.

Finally, if we want more flexibility, we can use different $\beta$ thresholds for the forward dropping steps and the backward moves. 

As for illustration, we have the following algorithm,

\begin{center}
	\fbox{%
		\begin{minipage}{4.7 in}
	\textbf{Dropping forward-backward algorithm with Mallows's $C_p$ for regression}	\begin{enumerate} 
		\item \textbf{Input}: a set of feature $C=\{X_1,X_2,...,X_p\}$, response $Y$, $\alpha$-to enter, $\beta$-to remove. 
		
		\item \textbf{Output}: a set $R\subset C$ of relevant features.
		
		\item \textbf{Forward-dropping steps:} Sequentially add to R the feature that minimizes $C_p$. Remove from C this feature and the features that can not reduce $C_p$ more than an amount of $\beta$. Stop when no feature can reduce $C_p$ more than $\alpha$.
		
		\item \textbf{Re-forward steps:}	
		\begin{enumerate}
			\item  $C=\{X_1,X_2,...,X_p\}\setminus R,$
			
			\item Sequentially add to R a feature that minimizes $C_p$ until no feature can reduce $C_p$ more than $\alpha$.
			
		\end{enumerate}
		\item \textbf{Backward steps:} Sequentially remove from R the least useful feature, one at a time, if removing that feature causes $C_p$ to increase $C_p$ no more than an amount $\beta$.
		
	\end{enumerate}
\end{minipage}}
\end{center}

\section{Experiments}\label{exper}
\subsection{Description}
In this section, we illustrate the power of our method by comparing the \textit{dropping forward-backward algorithm} to the stepwise algorithm and the intuitive forward-backward scheme mentioned in the last part of section \ref{review} on artificial and real data. Note that throughout all the experiments, we carry out standard normalization procedures for every dataset. 

For the simulation, we generate $n=80$ samples of dimension $p$, where $p$ varies from 50 to 80. The original regression model is $Y = 4.5+3X_1+2.1X_2+3.5X_7+ 0.8X_{12} + \epsilon$, where $\epsilon\sim \mathcal{N}(0,2)$. We repeat each experiment 1000 times for each value of $p$ and report the average error sum of squares and the average number of selected features. We choose $\alpha = \beta = 0.01$ and use $C_p$ as the selection criterion. The results are shown in table \ref{pvary}. We do not mention the regression error here, as they are very low and are the same when rounding off to five decimal places.

The experiments on real data are feature selection for classification based on \textit{trace criterion}. Trace criterion is popular class separability measure for feature selection in classification task (more details in \cite{fukunaga2013introduction},\cite{johnson2002applied}). There are many equivalent versions. However, suppose that we have $C$ classes, and there are $n_i$ observation for the $i^{th}$ class, then one way to define the criterion is
\begin{equation}
	trace(S_w^{-1}S_b),
\end{equation}
where 
\begin{equation}
S_b = \sum_{i=1}^Cn_i(\bar{x}_i-\bar{x})(\bar{x}_i-\bar{x})'
\end{equation}
and 
\begin{equation}
	S_w = \sum_{i=1}^C\sum_{j=1}^{n_i}( {x}_{ij}-\bar{x}_i)({x}_{ij}-\bar{x}_i)'
\end{equation}
are the between-class scatter matrix and within-class scatter matrix, respectively. Here, $\bar{x}_i$ is the mean for the $i^{{th}}$ class, $\bar{x}$ is the overall mean. 

Since this criterion measures the separability of classes, we would like to maximize it. After selecting the relevant features, we classify the samples using a \textit{support vector machine (SVM)} classifier and a \textit{linear discriminant analysis (LDA)} classifier and compare results among the feature selection methods and when all the features are used. For Parkinson, since the dimension highly exceeds the number of samples, we use Principal Component Analysis to extract the first 200 components that explain 98.58\%  of the variance, and then use SVM or LDA to classify samples. 

The information about the datasets from UCI repository that we use are summarized in table \ref{data}. For the datasets that do not have separate training, testing sets, we randomly split the data into training and testing set. \begin{table}
		\caption{UCI datasets for experiments}		\label{data}
	\begin{center}
		\begin{tabular}{|c|c|c|c|c|}
			\hline
			dataset   & \# classes & \# training & \# testing samples & \# dimensions \\ \hline
			Biodeg    & 2          & 1055        & (1.35,21.6)        & 41            \\ \hline
			Inosphere & 2          & 351         & (0.65,23.2)        & 34            \\ \hline
			Optic     & 10         & 3823        & 1797               & 64            \\ \hline
			Satellite & 6          & 4435        & 2000               & 36            \\ \hline
			Parkinson & 2          & 378         & 378                & 753           \\ \hline
		\end{tabular}
	\end{center}
\end{table}

\subsection{Results and discussion}

\begin{table}
		\caption{The performances of the three procedures on simulated data.}\label{pvary}	
	\begin{center}
		\begin{tabular}{|c|c|c|c|}
			\hline
			\multirow{2}{*}{Dimension} & \multicolumn{3}{c|}{\begin{tabular}[c]{@{}c@{}} (time in second(s), number of selected features)\end{tabular}} \\ \cline{2-4} 
			& Dropping forward-backward                       & Forward-backward                      & Stepwise                            \\ \hline
			$p=50$                     & (0.2576, 3.915)                                  & (0.3005, 3.914)                        & (0.3083, 3.914)                      \\ \hline
			$p=60$                     & (0.2965, 4.002)                                  & (0.3692, 4)                            & (0.3776, 4)                          \\ \hline
			$p=70$                     & (0.3437,4.001)                                  & (0.44,4)                             & (0.4470,4)                          \\ \hline
			$p=80$                     & (0.4140,4)                                      & (0.4937,4)                            & (0.5012,4)                          \\ \hline
		\end{tabular}
	\end{center}
\end{table}
From table \ref{pvary}, we see that dropping forward-backward procedure significantly surpasses the other two methods in term of speed (when $p=70$, the speed of dropping forward-backward procedure is 21.89\% less than forward-backward algorithm and 23.11\% less than stepwise algorithm), but rarely increases the number of features in the model (at most only twice in a thousand times when $p=60$ in this simulation study). 

\begin{table}
		\caption{The speed and number of selected features of the procedures on real data with $\alpha=\beta=0.05$, except for Parkinson, we use $\alpha=0.05, \beta = 0.01$ according to the discussion in section \ref{dfb}.}	
		\label{real}	
	\begin{center}		
		\begin{tabular}{|c|c|c|c|}
			\hline
			\multirow{2}{*}{Dataset} & \multicolumn{3}{c|}{(time (s), number of selected features)} \\ \cline{2-4} 
			& Dropping forward-backward     & Stepwise        & Forward-backward    \\ \hline
			
			Biodegradation                   & (0.1375,5)                  & (0.212, 6)    & (0.204, 6)       \\ \hline
			Ionosphere                & (0.258,14)                  & (0.437,14)     & (0.340,14)        \\ \hline
			Optic                    & (24.045,49)                    & (70.250,49)      & (36.282,49)          \\ \hline
			Satellite                & (2.801,17)                     & (3.595,14)       & (2.902,14)           \\ \hline
			Parkinson                & (2.691,10)                     & (25.486,24)       & (25.685,24)           \\ \hline				
		\end{tabular}
	\end{center}
\end{table}

\begin{figure}
	\centering
	\includegraphics[scale=0.7]{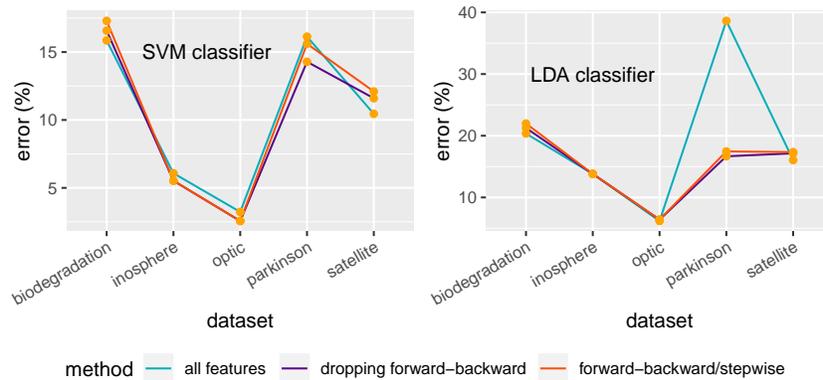}
	\caption{performances of three approaches on real data using LDA classifier. Note that stepwise and forward-backward selection give the same error rates, so we plot them on the same line.}
	\label{classi}
\end{figure}
For real data, we can see from table \ref{real} that dropping the forward-backward procedure highly surpasses the other two methods in terms of speed. Specifically, for the optic dataset, the speed of dropping forward backward is only 34.23\% the speed of stepwise procedure and only 66.29\% the speed of the forward-backward procedure. Also, when combining with figure \ref{classi}, we see that the dropping forward-backward approach has close performances, and many times, better than stepwise and forward-backward procedures, depending on the classifier. For biodegradation and satellite datasets, the features selection methods procedure subsets of features that can obtain a close error rate for using all features. However, multiple times, the three features selection methods reach lower error rates than using all the features available.    

Note that for Parkinson data set, the performances of all three feature selection methods highly surpass the PCA feature extraction approach, especially when using the LDA classifier. One possible explanation for this is that PCA suffers from the poor estimation of the covariance matrix. 

Another worth-noticing thing from table \ref{real} is that the amount of time it takes for the three feature selection procedures to terminate for Parkinson data set is far less than for Optic, even though Parkinson has 753 features and Optic has only 64. This implies that the time it takes for the procedures to run depends not only on the dimension of the data but also on the sparsity of the resulting model.
\section{Conclusion}\label{concl}
In this paper, we point out some issues with forward and stepwise feature selection and from that propose a new faster scheme that can maintain a good performance. We illustrate the power of our method via simulation and experiments on real datasets. We also give an example to show that for datasets where the number of features highly exceeds the number of samples, feature selection may be preferable, since feature extraction using PCA may suffer from the poor estimation of the covariance matrix.  Our examples also illustrated that the amount of time it takes for selection procedures to run depends not only on the dimension of the data but also on the sparsity of the resulting model.

Regarding how the algorithm should be implemented, we pointed out the choice of $\beta$ can play a crucial role in the speed of the algorithm and should be chosen according to the criterion used and the dimension of the dataset. Sometimes, the maximum number of features we would like to include may be much smaller than what the $\alpha,\beta$ thresholds produce. In such a case, we may specify the maximum number of features to be included. Note that we can also force the algorithm to print out the number of features included by forward moves, and after the forward dropping moves are the re-forward moves, allowing us to rank the features.  
  
\bibliography{bbli}

\begin{thebibliography}{10}
\providecommand{\url}[1]{\texttt{#1}}
\providecommand{\urlprefix}{URL }
\providecommand{\doi}[1]{https://doi.org/#1}

\bibitem{billings2007sparse}
Billings, S.A., Wei, H.L.: Sparse model identification using a forward
  orthogonal regression algorithm aided by mutual information. IEEE
  Transactions on Neural Networks  \textbf{18}(1),  306--310 (2007)

\bibitem{couvreur2000optimality}
Couvreur, C., Bresler, Y.: On the optimality of the backward greedy algorithm
  for the subset selection problem. SIAM Journal on Matrix Analysis and
  Applications  \textbf{21}(3),  797--808 (2000)

\bibitem{donoho2005stable}
Donoho, D.L., Elad, M., Temlyakov, V.N.: Stable recovery of sparse overcomplete
  representations in the presence of noise. IEEE Transactions on information
  theory  \textbf{52}(1),  6--18 (2005)

\bibitem{fukunaga2013introduction}
Fukunaga, K.: Introduction to statistical pattern recognition. Elsevier (2013)

\bibitem{johnson2002applied}
Johnson, R.A., Wichern, D.W., et~al.: Applied multivariate statistical
  analysis, vol.~5. Prentice hall Upper Saddle River, NJ (2002)

\bibitem{kumar2014feature}
Kumar, V., Minz, S.: Feature selection: a literature review. SmartCR
  \textbf{4}(3),  211--229 (2014)

\bibitem{liu2012feature}
Liu, H., Motoda, H.: Feature selection for knowledge discovery and data mining,
  vol.~454. Springer Science \& Business Media (2012)

\bibitem{mao2004orthogonal}
Mao, K.Z.: Orthogonal forward selection and backward elimination algorithms for
  feature subset selection. IEEE Transactions on Systems, Man, and Cybernetics,
  Part B (Cybernetics)  \textbf{34}(1),  629--634 (2004)

\bibitem{tropp2004greed}
Tropp, J.A.: Greed is good: Algorithmic results for sparse approximation. IEEE
  Transactions on Information theory  \textbf{50}(10),  2231--2242 (2004)

\bibitem{zhang2011adaptive}
Zhang, T.: Adaptive forward-backward greedy algorithm for learning sparse
  representations. IEEE transactions on information theory  \textbf{57}(7),
  4689--4708 (2011)

\end{thebibliography}
\bibliographystyle{splncs04}
\pagebreak
\appendix
\section{Appendix}
\textbf{Proof for theorem \ref{theorem}:}
Let $K=Z'Z$ and denote by $[U]_{rs}$ the $(r,s)$ entries of a matrix $U$. The proof of the above theorem follows from these remarks:

\begin{itemize}
	\item \textit{Remark 1:} The determinant of $X'X$ does not change if we interchange any columns of $X$, i.e., $|K|=|X'X|$. 
	
	\item \textit{Remark 2:}  $[K^{-1}]_{is} = [(X'X)^{-1}]_{js},[K^{-1}]_{js}=[(X'X)^{-1}]_{is}$ for $s\neq i,j$.
	
	\item   \textit{Remark 3:} $[K^{-1}]_{jj} = [(X'X)^{-1}]_{ii}, [K^{-1}]_{ii} = [(X'X)^{-1}]_{jj}$.
	\item   \textit{Remark 4:} $[K^{-1}]_{ij} = [(X'X)^{-1}]_{ji}, [K^{-1}]_{ji} = [(X'X)^{-1}]_{ij}$.
\end{itemize}

From remarks 1-4, we see that we can get $K^{-1}$ from  $(X'X)^{-1}$ by interchanging its $i^{th},j^{th}$ rows, and then, its $i^{th},j^{th}$ columns. Moreover, $\hat{\beta} = (X'X)^{-1}X'Y,\;\widehat{\gamma} = (Z'Z)^{-1}Z'Y$  and  

$$Z'Y=\begin{pmatrix}x_1'\\\vdots\\x_j'\\\vdots\\x_i'\\\vdots\\x_p'\end{pmatrix}Y = \begin{pmatrix}x_1'Y\\\vdots\\x_j'Y\\\vdots\\x_i'Y\\\vdots\\x_p'Y\end{pmatrix},$$
which implies that we can get $Z'Y$ from $X'Y$ by swapping its $i^{th},j^{th}$ entries. Hence, we can get $\widehat{\gamma}$  by swapping the $i^{th}, j^{th}$ position of $\hat{\beta}$, i.e.,
$$\widehat{\gamma} = (\hat{\beta}_1,...,\hat{\beta}_{i-1},\hat{\beta}_j,\hat{\beta}_{i+1},...,\hat{\beta}_{j-1},\hat{\beta}_i,\hat{\beta}_{j+1},...,\hat{\beta}_p)'.$$

\textbf{Proof of the remarks:}
\begin{itemize}
	\item Proof of \textit{remark 1:} 
	\begin{align}\label{xxone}
	K = \begin{pmatrix}x_1'\\\vdots\\x_j'\\\vdots\\x_i'\\x_p'\end{pmatrix} (x_1,...,x_j,...,x_i,...,x_p)
	= \begin{pmatrix}
	x_1'x_1&...&x_1'x_j&...&x_1'x_i&...&x_1'x_p\\
	\vdots & \ddots & \vdots & \ddots & \vdots & \ddots &\vdots\\
	x_j'x_1&...&x_j'x_j&...&x_j'x_i&...&x_j'x_p\\
	\vdots & \ddots & \vdots & \ddots & \vdots & \ddots &\vdots\\
	x_i'x_1&...&x_i'x_j&...&x_i'x_i&...&x_i'x_p\\
	\vdots & \ddots & \vdots & \ddots & \vdots & \ddots &\vdots\\
	x_p'x_1&...&x_p'x_j&...&x_p'x_i&...&x_p'x_p
	\end{pmatrix}
	\end{align}
	Moreover, 
	\begin{align}\label{xxtwo}
	X'X = \begin{pmatrix}x_1'\\\vdots\\x_i'\\\vdots\\x_j'\\\vdots\\x_p'\end{pmatrix} (x_1,...,x_i,...,x_j,...,x_p)
	= \begin{pmatrix}
	x_1'x_1&...&x_1'x_i&...&x_1'x_j&...&x_1'x_p\\
	\vdots & \ddots & \vdots & \ddots & \vdots & \ddots &\vdots\\
	x_i'x_1&...&x_i'x_i&...&x_i'x_j&...&x_i'x_p\\
	\vdots & \ddots & \vdots & \ddots & \vdots & \ddots &\vdots\\
	x_j'x_1&...&x_j'x_i&...&x_j'x_j&...&x_j'x_p\\
	\vdots & \ddots & \vdots & \ddots & \vdots & \ddots &\vdots\\
	x_p'x_1&...&x_p'x_i&...&x_p'x_j&...&x_p'x_p
	\end{pmatrix}.
	\end{align}
	Hence, the after interchanging two columns, we can get the new $X'X$ by interchange the $i^{th},j^{th}$ rows and then the  $i^{th},j^{th}$ columns of the original $X'X$. Therefore, their determinants are the same.
	%-------------------------------------------
	
	From \ref{xxone}, \ref{xxtwo}, we have $M_{ij}$, the determinant of the $(p-1) \times (p-1)$ matrix that results from deleting row i and column j of $K$, is equal to  $N_{ij}$, the determinant of the $(p-1) \times (p-1)$ matrix that results from deleting row i and column j of $X'X$. Moreover, from remark 1, we know that the determinant of $X'X$ does not change if we swap any columns of $X$, for $r\neq i,j$ and $s\neq i,j$. Therefore,
	$$[K^{-1}]_{rs} = \frac{(-1)^{r+s}M_{rs}}{|X'X|} = \frac{(-1)^{r+s}N_{rs}}{|X'X|} = [X'X]_{rs}.$$
	
	\item Proof of \textit{remark 2:} for $s\neq i,j$, 
	\begin{align*}
	[K^{-1}]_{is} = \frac{(-1)^{i+s}}{|X'X|} |A_{is}|,
	\end{align*}
	where 
	\begin{align*}
	A_{is}=\begin{pmatrix}
	x_1'x_1 &...  &x_1'x_{s-1}   &x_1'x_{s+1}  &...  &x_1'x_p \\ 
	\vdots & \ddots & \vdots  &\vdots  &\ddots  &\vdots \\ 
	x_{i-1}'x_1 & ... & x_{i-1}'x_{s-1}  &x_{i-1}'x_{s+1}  &...  & x_{i-1}'x_p \\ 
	x_{i+1}'x_1 & ... & x_{i+1}'x_{s-1}   &x_{i+1}'x_{s+1}  &...  & x_{i+1}'x_p \\ 
	\vdots & \ddots & \vdots   &\vdots  &\ddots  &\vdots \\ 
	x_i'x_1 &...  &x_i'x_{s-1}   &x_i'x_{s+1}  &...  &x_i'x_p \\ 
	\vdots & \ddots & \vdots  &\vdots  &\ddots  &\vdots \\ 
	x_p'x_1 &...  &x_p'x_{s-1}  &x_p'x_{s+1}  &...  &x_p'x_p \\ 
	\end{pmatrix}.
	\end{align*}
	Moreover, 
	\begin{align*}
	[(X'X)^{-1}]_{js} = \frac{(-1)^{j+s}}{|X'X|} |B_{js}|,
	\end{align*}
	where
	\begin{align*}
	B_{js}=\begin{pmatrix}
	x_1'x_1 &...  &x_1'x_{s-1} &x_1'x_{s+1}  &...  &x_1'x_p \\ 
	\vdots & \ddots & \vdots  &\vdots  &\ddots  &\vdots \\ 
	x_{i}'x_1 & ... & x_i'x_{s-1}  &x_i'x_{s+1}  &...  & x_i'x_p \\ 
	\vdots & \ddots & \vdots &\vdots  &\ddots  &\vdots \\ 
	x_{j-1}'x_1 & ... & x_{j-1}'x_{s-1}   &x_{j-1}'x_{s+1}  &...  & x_{j-1}'x_p \\ 
	x_{j+1}'x_1 & ... & x_{j+1}'x_{s-1}  &x_{j+1}'x_{s+1}  &...  & x_{j+1}'x_p \\ 
	\vdots & \ddots & \vdots &\vdots  &\ddots  &\vdots \\ 
	x_p'x_1 &...  &x_p'x_{s-1}   &x_p'x_{s+1}  &...  &x_p'x_p \\ 
	\end{pmatrix}.
	\end{align*}
	Note that we can get $B_{js}$ from $A_{is}$ by doing the following swaps:
	
	$\;\;\;\;\;\;\;\;\;(j-1)^{th}$ row $\leftrightarrow (j-2)^{th}$ row,
	
	$\;\;\;\;\;\;\;\;\;(j-2)^{th}$ row $\leftrightarrow (j-3)^{th}$ row,
	
	$\;\;\;\;\;\;\;\;\;\;\;\;\;\;\;\;\;\;\;\;\;\;\;\;\vdots$
	
	$\;\;\;\;\;\;\;\;\;(i+1)^{th}$ row $\leftrightarrow i^{th}$ row,
	
	and then swap the original $i^{th}, j^{th}$ columns. Hence, we made $(j-i-1)+1$ swaps. Therefore,
	$$[K^{-1}]_{is} = (-1)^{i+s} (-1)^{j-i-1+1} |B_{js}|=(-1)^{j+s}|B_{js}|=[(X'X)^{-1}]_{js}.$$
	
	Similarly, we can prove that $[K^{-1}]_{js}=[(X'X)^{-1}]_{is}$ for $s\neq i,j$.
	\item Proof of \textit{remark 3:} 
	\begin{align*}
	[K^{-1}]_{ii}=\frac{(-1)^{i+i}|A_{ii}|}{|X'X|}
	\end{align*}
	where \begin{align*}
	A_{ii}=\begin{pmatrix}
	x_1'x_1 &...  &x_1'x_{i-1}   &x_1'x_{i+1}  &...&x_1'x_i&...  &x_1'x_p \\ 
	\vdots & \ddots & \vdots  &\vdots &\ddots&\vdots &\ddots  &\vdots \\ 
	x_{i-1}'x_1 & ... & x_{i-1}'x_{i-1}  &x_{i-1}'x_{i+1}  &...  &x_{i-1}'x_i&...& x_{i-1}'x_p \\ 
	x_{i+1}'x_1 & ... & x_{i+1}'x_{i-1}   &x_{i+1}'x_{i+1}  &... &x_{+-1}'x_i&... & x_{i+1}'x_p \\ 
	\vdots & \ddots & \vdots  &\vdots &\ddots&\vdots &\ddots  &\vdots \\ 
	x_i'x_1 &...  &x_i'x_{i-1}   &x_i'x_{i+1}  &... &x_i'x_i&... &x_i'x_p \\ 
	\vdots & \ddots & \vdots  &\vdots &\ddots&\vdots &\ddots  &\vdots \\ 
	x_p'x_1 &...  &x_p'x_{i-1}  &x_p'x_{i+1}  &...&x_p'x_i&...  &x_p'x_p 
	\end{pmatrix}
	\end{align*}
	Note that 
	\begin{align*}
	[(X'X)^{-1}]_{jj}=\frac{(-1)^{j+j}|B_{jj}|}{|X'X|},
	\end{align*}
	where \begin{align*}
	B_{jj}=\begin{pmatrix}
	x_1'x_1 &... &x_1'x_i&... &x_1'x_{j-1}   &x_1'x_{j+1}  &...  &x_1'x_p \\ 
	\vdots & \ddots &\vdots & \ddots & \vdots  &\vdots &\ddots&\vdots \\ 
	x_i'x_1 &... &x_i'x_i&... &x_i'x_{j-1}   &x_i'x_{j+1}  &...  &x_i'x_p \\ 
	\vdots & \ddots &\vdots & \ddots & \vdots  &\vdots &\ddots&\vdots s \\ 
	x_{j-1}'x_1 & ... & x_{j-1}'x_i  &... &x_{j-1}'x_{j-1}  &x_{j-1}'x_{j+1}&... & x_{j-1}'x_p \\ 
	x_{j+1}'x_1 & ... & x_{j+1}'x_i  &... &x_{j+1}'x_{j-1}  &x_{j+1}'x_{j+1}&... & x_{j+1}'x_p \\ 
	\vdots & \ddots &\vdots & \ddots & \vdots  &\vdots &\ddots&\vdots  \\ 
	x_p'x_1&... &x_p'x_i &...  &x_p'x_{j-1}  &x_p'x_{j+1}  &... &x_p'x_p.
	\end{pmatrix}.
	\end{align*}
	We can get $B_{jj}$ from $A_{ii}$ by doing the following swaps:
	
	$\;\;\;\;\;\;\;\;\;(j-1)^{th}$ row $\leftrightarrow (j-2)^{th}$ row,
	
	$\;\;\;\;\;\;\;\;\;\;\;\;\;\;\;\;\;\;\;\;\;\;\;\;\vdots$
	
	$\;\;\;\;\;\;\;\;\;(i+1)^{th}$ row $\leftrightarrow i^{th}$ row,
	
	and then,
	
	$\;\;\;\;\;\;\;\;\;(j-1)^{th}$ column $\leftrightarrow (j-2)^{th}$ column,
	
	$\;\;\;\;\;\;\;\;\;\;\;\;\;\;\;\;\;\;\;\;\;\;\;\;\vdots$
	
	$\;\;\;\;\;\;\;\;\;(i+1)^{th}$ column $\leftrightarrow i^{th}$ column. 
	
	Hence, we made $2(j-i-1)$ swaps. Therefore, $|A_{ii}|=|B_{jj}|$, which implies, 
	$$[K^{-1}]_{ii} = [(X'X)^{-1}]_{jj}.$$
	Similarly, we can prove that 
	$$[K^{-1}]_{jj} = [(X'X)^{-1}]_{ii}.$$
	
	\item Proof of \textit{remark 4:} 
	
	$$[K^{-1}]_{ij}=\frac{(-1)^{i+j}}{|X'X|}|A_{ij}|,$$
	where $$A_{ij}=\begin{pmatrix}
	x_1'x_1&...&x_1'x_j&...&x_1'x_{j-1}&x_1'x_{j+1}&...&x_1'x_p\\
	\vdots&\ddots&\vdots&\ddots&\vdots&\vdots&\ddots&\vdots\\
	x_{i-1}'x_1&...&x_{i-1}'x_j&...&x_{i-1}'x_{j-1}&x_{i-1}'x_{j+1}&...&x_{i-1}'x_p\\
	x_{i+1}'x_1&...&x_{i+1}'x_j&...&x_{i+1}'x_{j-1}&x_{i+1}'x_{j+1}&...&x_{i+1}'x_p\\
	\vdots&\ddots&\vdots&\ddots&\vdots&\vdots&\ddots&\vdots\\
	x_i'x_1&...&x_i'x_j&...&x_i'x_{j-1}&x_i'x_{j+1}&...&x_i'x_p\\
	\vdots&\ddots&\vdots&\ddots&\vdots&\vdots&\ddots&\vdots\\
	x_p'x_1&...&x_p'x_j&...&x_p'x_{j-1}&x_p'x_{j+1}&...&x_p'x_p
	\end{pmatrix}.$$
	Moreover, 
	$$[(X'X)^{-1}]_{ji}=\frac{(-1)^{i+j}}{|X'X|}|B_{ji}|,$$
	where 
	$$B_{ji}=\begin{pmatrix}
	x_1'x_1&...&x_1'x_{i-1}&x_1'x_{i+1}&...&x_1'x_j&...&x_1'x_p\\
	\vdots&\ddots&\vdots&\vdots&\ddots&\vdots&\ddots&\vdots\\
	x_i'x_1&...&x_i'x_{i-1}&x_i'x_{i+1}&...&x_i'x_j&...&x_i'x_p\\
	\vdots&\ddots&\vdots&\vdots&\ddots&\vdots&\ddots&\vdots\\
	x_{j-1}'x_1&...&x_{j-1}'x_{i-1}&x_{j-1}'x_{i+1}&...&x_{j-1}'x_j&...&x_{j-1}'x_p\\
	x_{j+1}'x_1&...&x_{j+1}'x_{i-1}&x_{j+1}'x_{i+1}&...&x_{j-1}'x_j&...&x_{j+1}'x_p\\
	\vdots&\ddots&\vdots&\vdots&\ddots&\vdots&\ddots&\vdots\\
	x_p'x_1&...&x_p'x_{i-1}&x_p'x_{i+1}&...&x_p'x_j&...&x_p'x_p
	\end{pmatrix}.$$
	Note that we can get $B_{ji}$ by doing the following swaps:
	
	$\;\;\;\;\;\;\;\;\;(j-1)^{th}$ row $\leftrightarrow (j-2)^{th}$ row,
	
	$\;\;\;\;\;\;\;\;\;\;\;\;\;\;\;\;\;\;\;\;\;\;\;\;\vdots$
	
	$\;\;\;\;\;\;\;\;\;(i+1)^{th}$ row $\leftrightarrow i^{th}$ row,
	
	and then,
	
	$\;\;\;\;\;\;\;\;\;(i+1)^{th}$ column $\leftrightarrow (i+2)^{th}$ column,
	
	$\;\;\;\;\;\;\;\;\;\;\;\;\;\;\;\;\;\;\;\;\;\;\;\;\vdots$
	
	$\;\;\;\;\;\;\;\;\;(j-2)^{th}$ column $\leftrightarrow (j-1)^{th}$ column. 
	
	Hence, we made $2(j-i-1)$ swaps. Therefore, $|A_{ij}|=|B_{ji}|$, which implies, 
	$$[K^{-1}]_{ij} = [(X'X)^{-1}]_{ji}.$$
	Similarly, we can prove that 
	$$[K^{-1}]_{ji} = [(X'X)^{-1}]_{ij}.$$
\end{itemize}
\end{document}